\documentclass[journal]{IEEEtran}
\usepackage[utf8]{inputenc}

\usepackage[english]{babel}
\usepackage{amsmath, amssymb}
\usepackage{float}
\usepackage{makecell}
\usepackage{multirow}
\usepackage{cite}
\usepackage{amsmath,amssymb,amsfonts}
\usepackage{algorithmicx}
\usepackage{amsmath}
\usepackage{amsfonts}
\usepackage{amssymb}\usepackage{graphicx}
\usepackage{textcomp}
\usepackage{xcolor}

\usepackage{subfigure}
\usepackage{epstopdf}
\usepackage{aecompl}
\usepackage{times}
\usepackage{epsfig}
\usepackage{graphicx}
\usepackage[T1]{fontenc}    
\usepackage{url}            
\usepackage{booktabs}       
\usepackage{nicefrac}       
\usepackage[ruled,vlined]{algorithm2e}

\ifCLASSINFOpdf
 
\else
 
\fi

\begin{document}
%

\title{Residual 3D Scene Flow Learning with Context-Aware Feature Extraction}

\author{Guangming Wang, Yunzhe Hu, Xinrui Wu, and Hesheng Wang
	\thanks{This work was supported in part by the Natural Science Foundation of China under Grant 62073222 and U1913204, in part by “Shu Guang”project supported by Shanghai Municipal Education Commission and Shanghai Education Development Foundation under Grant 19SG08,  in part by Shenzhen Science and Technology Program under Grant JSGG20201103094400002, in part by the Science and Technology Commission of Shanghai Municipality under Grant 21511101900, in part by grants from NVIDIA Corporation. Corresponding Author: Hesheng Wang. The first two authors contributed equally.}
	\thanks{G. Wang, Y. Hu, X. Wu, and H. Wang are with Department of Automation, Key Laboratory of System Control and Information Processing of Ministry of Education, Key Laboratory of Marine Intelligent Equipment and System of Ministry of Education, Shanghai Engineering Research Center of Intelligent Control and Management, Shanghai Jiao Tong University, Shanghai 200240, China.}}

\markboth{Journal of \LaTeX\ Class Files,~Vol.~14, No.~8, August~2015}%
{Shell \MakeLowercase{\textit{et al.}}: Bare Demo of IEEEtran.cls for IEEE Journals}

\maketitle

\begin{abstract}
    Scene flow estimation is the task to predict the point-wise or pixel-wise 3D displacement vector between two consecutive frames of point clouds or images, which has important application in fields such as service robots and autonomous driving. Although many previous works have explored greatly on scene flow estimation based on point clouds, there are two problems that have not been noticed or well solved before: 1) Points of adjacent frames in repetitive patterns may be wrongly associated due to similar spatial structure in their neighbourhoods; 2) Scene flow between adjacent frames of point clouds with long-distance movement may be inaccurately estimated. To solve the first problem, a novel context-aware set convolution layer is proposed in this paper to exploit contextual structure information of Euclidean space and learn soft aggregation weights for local point features. This design is inspired by human perception of contextual structure information during scene understanding with repetitive patterns. The context-aware set convolution layer is incorporated in a context-aware point feature pyramid module of 3D point clouds for scene flow estimation. For the second problem, an explicit residual flow learning structure is proposed in the residual flow refinement layer to cope with long-distance movement. The experiments and ablation study on FlyingThings3D and KITTI scene flow datasets demonstrate the effectiveness of each proposed component. The qualitative results show that the problems of ambiguous inter-frame association and long-distance movement estimation are well handled. Quantitative results on both FlyingThings3D and KITTI scene flow datasets show that the proposed method achieves state-of-the-art performance, surpassing all other previous works to the best of our knowledge by at least 25\%.  
\end{abstract}


%
\IEEEpeerreviewmaketitle

\section{Introduction}
Scene flow represents a 3D vector field that comprises point-wise or pixel-wise motion between two consecutive frames of point clouds or images. It provides a low-level and fundamental understanding of the motion of objects in a dynamic scene. The low-level information of scene flow can benefit many other tasks such as semantic segmentation \cite{wang2021anchor}, dynamic object segmentation \cite{9527213}, LiDAR odometry \cite{wang2021pwclo}, point cloud registration \cite{liu2019flownet3d}, etc.

Traditional works \cite{vogel20153d,menze2015object,wedel2011stereoscopic} mainly focus on acquiring optical flow and disparity from stereo or RGB-D images and extend the optical flow estimation to scene flow estimation. However, the performance of these methods is limited due to indirect optimization in the 3D environment. With the advancement of deep learning to process raw point clouds \cite{qi2017pointnet,qi2017pointnet++} and the application of LiDAR, recent studies \cite{liu2019flownet3d,gu2019hplflownet,wu2020pointpwc,wang2021hierarchical} focus on estimating scene flow with two consecutive frames of raw 3D point clouds as inputs. These methods predict scene flow using only 3D coordinates as inputs in an end-to-end fashion and do not require any prior knowledge of scene structure. FLowNet3D \cite{liu2019flownet3d} integrates set convolution layers based on PointNet++ \cite{qi2017pointnet++} to downsample input point clouds and learn deep hierarchical features, then introduces a flow embedding layer to associate points from their geometric similarities and encode the motions. The extracted flow embedding features are to be propagated through the set upconvolution layers to generate scene flow. HPLFlowNet \cite{gu2019hplflownet} proposes CorrBCLs to learn the correlation between two point clouds. However, due to the interpolation on permutohedral lattice, information of point clouds is inevitably lost therefore limiting the network performance. PointPWC-Net \cite{wu2020pointpwc} utilizes a patch-to-patch method to enlarge the receptive field of points while associating point clouds in cost volume layer. It is the first network to explore coarse-to-fine fashion on 3D point clouds. However, PointPWC-Net only uses the relative coordinates to learn the weights of points for flow embedding. HALFlow \cite{wang2021hierarchical} points out that correlation weights are not only decided by Euclidean space but by feature space. It thus proposes a novel double attentive embedding layer with two stages of attention in flow embedding.

However, all of the aforementioned scene flow estimation methods do not tackle to problem of recognizing repetitive patterns or well solve inaccurate long-distance motion estimation in a dynamic environment. For example, when estimating the motion of a multi-layer bookshelf, the points on the first layer of the shelf may be wrongly associated with those of the adjacent frame on the second layer because both layers share similar spatial structure information and are close in 3D space. In addition, for a fast-moving car, accurate scene flow predictions may not be made because the distance between two frames of point clouds is too long.

To tackle these two practical challenges, we propose a novel context-aware set convolution layer and a residual flow learning structure for the learning of 3D scene flow. For the first challenge of recognizing repetitive patterns, the key is to diminish the mismatch among the points in repetitive patterns. We argue that if points on a specific layer of a bookshelf store the information of which layer they belong to, then they can match the correct points on the same layer during the inter-frame association of two point clouds. Therefore, we propose a context-aware set convolution layer to enhance the recognition of relative spatial position of points in repetitive patterns during feature extraction. For the second challenge of long-distance motion estimation, PointPWC-Net \cite{wu2020pointpwc} introduces a coarse-to-fine scene flow estimation method. It estimates residual flow embedding features after obtaining coarse dense flow embedding features, then computes the overall flow embedding features. However, residual flow embedding features will bring about fuzziness in motion. We notice that scene flow can be directly added in 3D space, so we propose to explicitly learn the residual scene flow instead of flow embedding features. The learned residual flow will be directly added to the interpolated coarse dense flow and obtain a refined scene flow. The explicit estimation of residual scene flow will compensate directly for the long-distance motion estimation.

Our model is trained and evaluated on FlyingThings3D \cite{mayer2016large} dataset with synthetic data. We also evaluate our model without fine-tuning on KITTI Scene Flow \cite{menze2018object} dataset with real-world LiDAR scans to demonstrate its generalization ability. Our main contributions are as follows:
\begin{itemize}
	
	\item A novel context-aware set convolution layer is proposed to improve the recognition of repetitive patterns in 3D space. Contextual structure information in Euclidean space is exploited to learn the soft aggregation weights for the feature extraction of local points, which allows the extracted points to recognize their global position in a repetitive structure. 
	
	\item An explicit residual flow learning structure is proposed to compensate for long-distance motion. The learned residual flow is directly added to the coarse scene flow, which helps correct the estimated scene flow and eliminates the ambiguity from residual flow embedding features. 
		
	\item Effectiveness of the proposed context-aware set convolution layer and residual flow learning structure is demonstrated by ablation study. Quantitative results show that our method outperforms all the prior works by at least 25\% on FlyingThings3D and KITTI Scene Flow datasets. Visualization of certain objects with repetitive patterns and long-distance movement demonstrates that the challenges brought by these two situations are effectively handled.
\end{itemize}

\begin{figure*}[t]
	\centering
	\includegraphics[width=1.00\linewidth]{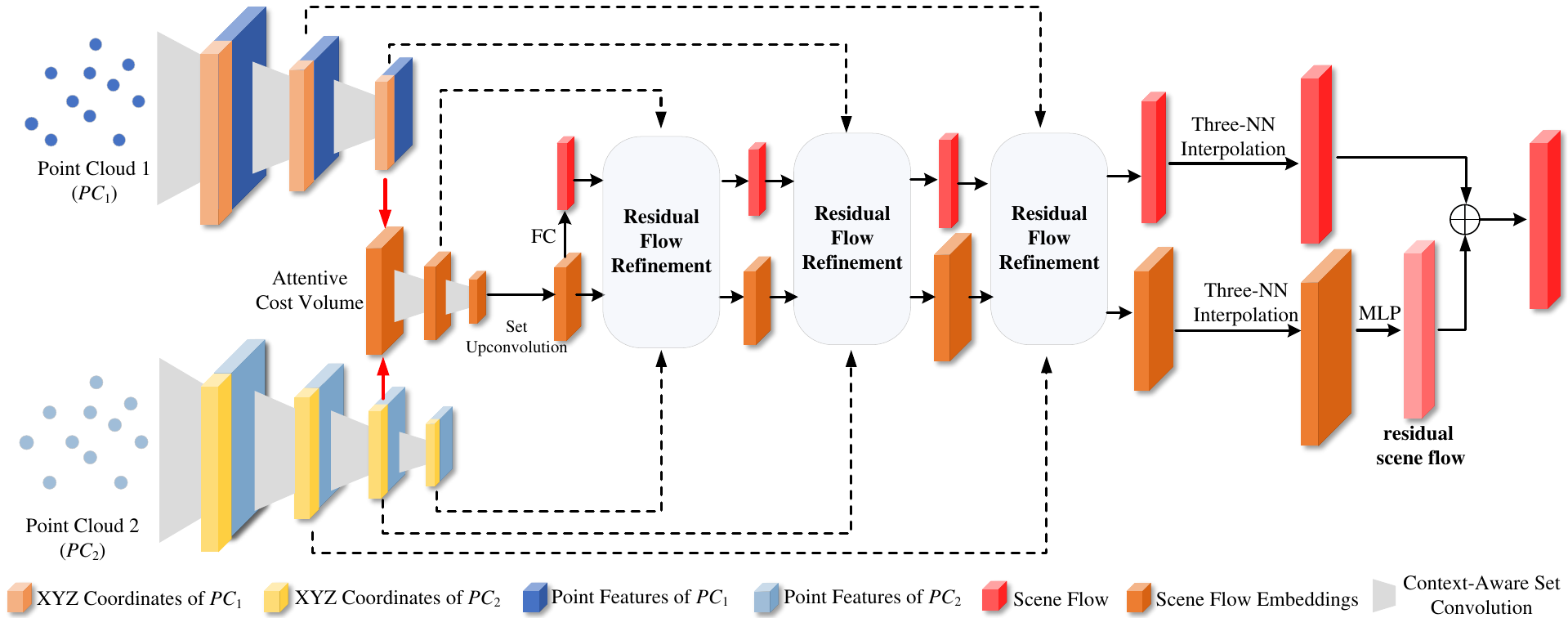}
	\vspace{-0.6cm}
	\caption{The details of our network architecture. There are three proposed context-aware set convolution layers for $PC_1$ and four for $PC_2$. The same level of layers shares the same parameters. The attentive cost volume \cite{wang2021hierarchical} is adopted to learn scene flow embedding features. Three proposed residual flow refinement layers are used in the flow refinement process. The scene flow between two points clouds are generated from coarse to fine.}   
	\label{figure:architecture}
\end{figure*}

\section{Related Work}
\label{section:related work}

\subsection{Deep Learning On 3D Data}
With the proposal of the pioneering work PointNet \cite{qi2017pointnet} and PointNet++ \cite{qi2017pointnet++}, deep learning on raw point clouds has gained more and more attention than voxels \cite{maturana2015voxnet,riegler2017octnet} and multi-view \cite{su2015multi,kalogerakis20173d} based methods. PointNet \cite{qi2017pointnet} is the first network  to process unstructured point sets and it utilizes shared Multi-Layer Perceptron (MLP) and max-pooling to learn the global features of point clouds. Follow-up work PointNet++ \cite{qi2017pointnet++} introduces a hierarchical fashion to learn local point features from neighbourhood points. SPLATNet \cite{su2018splatnet} maps input points to a high-dimensional permutohedral lattice and adopts Bilateral Convolution Layer (BCL) \cite{jampani2016learning} to perform convolutions on sparse lattice and smoothly interpolate filtered signals back to input points. DGCNN \cite{wang2019dynamic} builds dynamic graphs based on KNN query in the feature space and performs contextual feature aggregation. RandLA-Net \cite{hu2020randla} introduces attentive pooling to replace max-pooling in \cite{qi2017pointnet++} and effectively preserves useful local features from a wide neighbourhood. SOE-Net \cite{xia2021soe} designs a point orientation-encoding unit to integrate local information from eight orientations divided by eight octants.

\subsection{Point Cloud Based Scene Flow Estimation}

With the application of advanced LiDAR in autonomous driving and service robots, raw point clouds are becoming more accessible. Since the pioneering work of PointNet \cite{qi2017pointnet}, multiple deep learning methods have been utilized on raw point clouds for 3D scene flow estimation tasks.

Liu $et~al.$ \cite{liu2019flownet3d} propose FLowNet3D to learn scene flow end-to-end based on PointNet++  \cite{qi2017pointnet++}. FlowNet3D uses set conv layer following the architecture from \cite{qi2017pointnet++} to extract local features. Then, it uses one flow embedding layer to encode the motion between point clouds. A learnable set upconv layer is introduced to decode the flow embedding features and propagate coarse scene flow to a finer level to obtain the overall scene flow. Gu $et~al.$ \cite{gu2019hplflownet} introduce HPLFlowNet by leveraging Bilateral Convolutional Layers (BCL) and propose three novel layer designs: DownBCL, UpBCL, and CorrBCL. HPLFlowNet interpolates signals from input point clouds onto a permutohedral lattice and conducts sparse convolution on the lattice for scene flow estimation. FLOT \cite{puy2020flot} introduces an optimal transport module inspired by graph matching to learn the correspondences between two frames of point clouds. Wu $et~al.$ \cite{wu2020pointpwc} present PointPWC-Net, which follows a coarse-to-fine style for scene flow estimation. PointPWC-Net adopts the cost volume structure extensively used in optical flow estimation \cite{sun2018pwc} but extends it to a patch-to-patch manner. It uses MLP to learn the weights in aggregating the costs from different patches in the point clouds from adjacent frames. To improve the performance of cost volume, Wang $et~al.$ \cite{wang2021hierarchical} propose a novel double attentive flow embedding structure. It softly weighs the neighbourhood point features to associate adjacent frames and allocates more attention on the regions with correct correspondences. HCRF-Flow \cite{li2021hcrf} combines the strengths of Deep Neural Networks (DNNs) and Conditional Random Fields (CRFs) to estimate scene flow and proposes a continuous high-order CRFs module to model the spatial smoothness and rigid motion constraints for the refinement of point-wise predictions. FlowStep3D \cite{kittenplon2021flowstep3d} uses a global correction unit for all-to-all correspondence to generate the initial scene flow, and then uses a GRU-based recurrent architecture and a local update unit to refine the scene flow. The above-mentioned methods do not well solve the problem of recognizing repetitive patterns or long-distance movements. Our paper focuses on solving these issues and proposes a novel context-aware set convolution layer and residual flow learning structure to greatly improve the accuracy of scene flow estimation.

\section{Context-Aware Residual Flow Learning}
\label{section:method}

\subsection{Network Architecture}
\label{section:Network Architecture}

\begin{figure*}[t]
	\centering
	\includegraphics[width=1.00\linewidth]{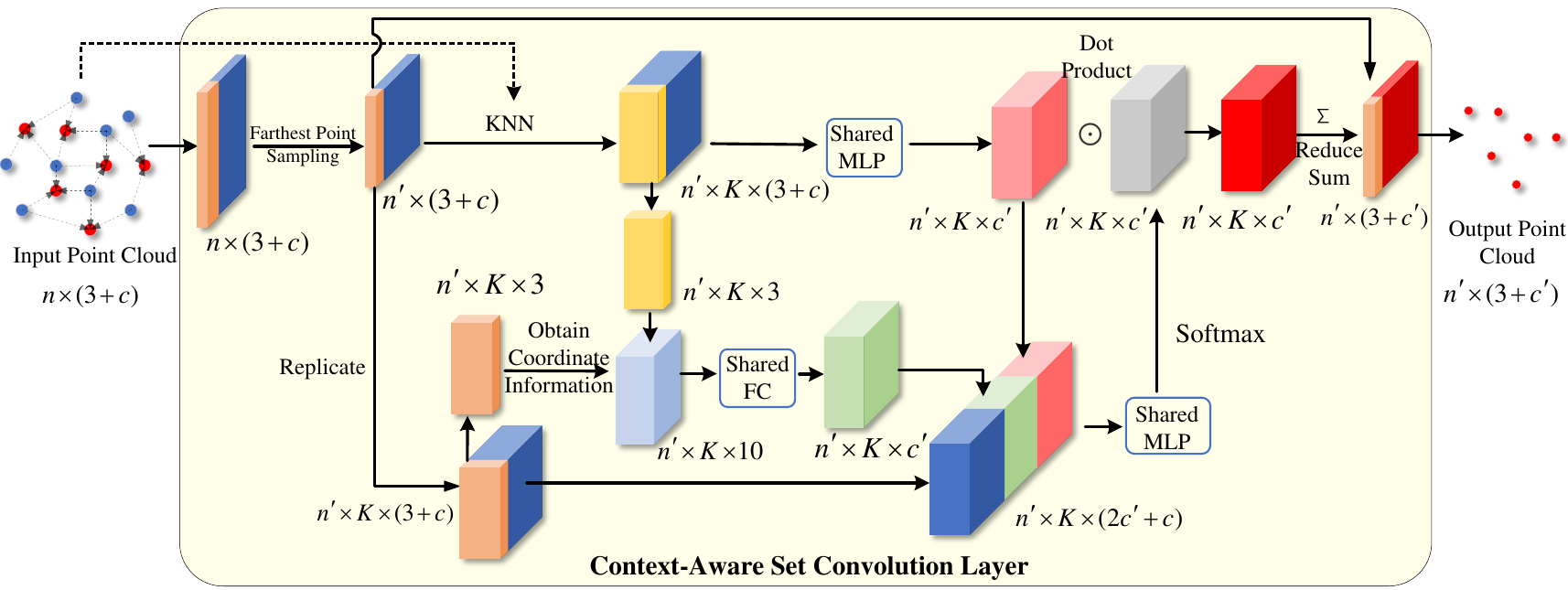}
	\vspace{-0.6cm}
	\caption{The illustration of our novel context-aware set convolution layer. Detailed process is elaborated in Section \ref{section:context-aware Set Conv Layer}.}   
	\label{figure:setconv}
\end{figure*}

Our complete network architecture is illustrated in Fig.~\ref{figure:architecture}. Given two consecutive frames of point cloud scans, i.e. $PC_1$ and $PC_2$, the proposed network predicts the 3D scene flow in a coarse-to-fine manner. Three main modules constitute our network: 1) Context-Aware Point Feature Pyramid, 2) Attentive Cost Volume, and 3) Hierarchical Residual Flow Refinement. The context-aware point feature pyramid module consists of three novel context-aware set convolution layers for $PC_1$ and four layers for $PC_2$, with each set convolution layer applying downsampling operation and aggregating local point features attentively as shown in Fig.~\ref{figure:setconv}. The pyramid feature encoding structure for each point cloud shares the same weights. The same cost volume module is adopted from \cite{wang2021hierarchical} followed by two set convolution layers. Then, upsampled flow embedding features are obtained from the lowest level of flow embedding features by the set upconvolution layer \cite{liu2019flownet3d}. A coarse scene flow is first estimated by a simple shared Fully Connection (FC) layer on the upsampled flow embedding features. Next, the coarse scene flow and the upsampled flow embedding features are passed into three residual flow refinement layers to be refined iteratively and hierarchically. We last employ Three Nearest Neighbours (Three-NN) interpolation and MLP to attain the final residual scene flow and interpolated scene flow. The final estimated scene flow will be the summation of the final residual flow and the interpolated scene flow.

\subsection{Context-Aware Point Feature Pyramid}
\label{section:context-aware Set Conv Layer}

In the context-aware point feature pyramid module, each point cloud is encoded through our novel context-aware set convolution layer hierarchically. Unlike the set convolution layer in \cite{qi2017pointnet++} which simply uses MLP and max-pooling to extract local features for each point in the scene, our context-aware set convolution layer employs the attention mechanism to avoid the point features lost in max-pooling operation. 

In order to retain the features information of all the neighbouring points in the raw point sets, a means of weighted feature extraction from all neighbouring points is proposed. For the purpose of aggregating local point features and exploiting the contextual structure information in 3D Euclidean space, we adopt the information of coordinates and features to obtain weights for nearest neighbours of each sampled points. The whole process is shown in Fig.~\ref{figure:setconv}.

For each context-aware set convolution layer, it takes $n$ points $\{p_i = \{x_i,pf_i\} ~|~i = 1,\dots,n\}$ as input, where $x_i \in \mathbb{R}^3$ denotes 3D coordinates and $pf_i \in \mathbb{R}^c$ denotes local point features. The output is $n' ~(n'<n)$ sampled and encoded points $\{p_j' = \{x_j',pf_j'\} ~|~j = 1,\dots,n'\}$, where $x_j' \in \mathbb{R}^3$ denotes 3D coordinates and $pf_j' \in \mathbb{R}^{c'}$ denotes local point features. In the very first context-aware set convolution layer of point feature pyramid, $pf_i$ is equal to $x_i$. For $n'$ output points $p_j'$, the 3D coordinates $x_j'$ are randomly sampled in the first layer and sampled by Farthest Point Sampling (FPS) algorithm in the subsequent layers, like \cite{wang2021hierarchical}. The extraction process of point features $pf_j'$ is elaborated as follows.

For each sampled point $p_j'$, its $K$ nearest neighbours are searched and grouped from the unsampled $n$ points as $\{p_j^k = \{x_j^k,pf_j^k\} ~|~k = 1,\dots,K\}$. Then, a learnable shared MLP will be applied to extract the features $pf_j'^k \in \mathbb{R}^{c}$ for the $K$ neighbouring points $p_j^k$ as follows:
\begin{equation}
	\label{eq:local features}
	pf_j'^k = MLP((x_j^k-x_j')\oplus pf_j^k),
\end{equation}
in which $\oplus$ denotes concatenation operation of two vectors.

Instead of adopting max-pooling operation to aggregate local features for $p_j'$, we will exploit and leverage spatial structure information and feature information to regress soft aggregation weights. We first use $x_j'$ and $x_j^k$ to obtain 3D spatial structure information as follows:
\begin{equation}
	\label{eq:coordinate info}
	d_j^k = x_j'\oplus x_j^k \oplus (x_j^k-x_j') \oplus ||x_j^k-x_j'||,
\end{equation}
where $||\cdot||$ indicates the $L_2$ norm. The contextual aggregation weights are calculated as:
\begin{equation}
	\label{eq:context-aware weight}
	w_j^k = \mathop{softmax}_{k=1,\dots,K}(MLP(FC(d_j^k)\oplus pf_j'^k \oplus pf_j)).
\end{equation}
In the calculation of contextual aggregation weights, the information of coordinates and relative position in 3D space of central point and its neighbours together with their features are utilized, which aims to encode the global position of central point in repetitive patterns.

Finally, the contextual aggregation weights are used to softly aggregate the neighbouring features for $p_j'$:
\begin{equation}
	\label{eq:aggregation}
	pf_j' = \sum_{k=1}^{K} pf_j'^k \odot w_j^k,
\end{equation}
where $\odot$ indicates dot product.

\subsection{Attentive Cost Volume}
\label{section:Cost Volume}
In order to learn the underlying motion information between two sets of points, we directly adopt the attentive cost volume in \cite{wang2021hierarchical} to associate two point clouds. The attentive cost volume module produces flow embedding features from the output of the context-aware point feature pyramid module for the flow estimation and refinement in the later process.

For point cloud $PC_1 = \{(x_i,f_i)~|~x_i \in \mathbb{R}^{3}, f_i \in \mathbb{R}^{c}, i = 1,\dots,n_1\}$ and point cloud $PC_2 = \{(y_j,g_j)~|~y_j \in \mathbb{R}^{3}, g_j \in \mathbb{R}^{c}, i = 1,\dots,n_2\}$, the flow embedding features $E = \{(x_i,e_i)~|~x_i \in \mathbb{R}^{3}, e_i \in \mathbb{R}^{c'}, i = 1,\dots,n_1\}$ between two point clouds are learned from attentive cost volume.

\subsection{Hierarchical Residual Flow Refinement}
\label{section:refinement}

The hierarchical residual flow refinement module mainly contains three residual flow refinement layers. The details of residual flow refinement layers are illustrated in Fig.~\ref{figure:refinement}. It contains four major components: 1) Set Upconvolution Layer, 2) Coordinates Warping Layers, 3) Attentive Cost Volume Layer, and 4) Scene Flow Predictor.

\subsubsection{Set Upconvolution Layer}
\label{set upconv}

We adopt set upconv layer in \cite{liu2019flownet3d} here to upsample coarse sparse flow embedding features. The inputs are $n$ sparse points with sparse flow embedding features $\{(x_i,e_i)|~ i=1,\dots,n \}$ and $n' ~(n'>n)$ dense points $\{(x_j',pf_j')|~ j=1,\dots,n' \}$ from the previous context-aware set convolution layer. The outputs are $n'$ dense flow embedding features $\{(x_j',e_j')|~ j=1,\dots,n' \}$. To be specific, each of the dense points will search and select its $K$ nearest neighbours in the sparse point sets, and the sparse flow embedding features will be aggregated into the dense flow embedding features in a learnable manner by shared MLP. 
\subsubsection{Coordinates Warping Layer}
\label{wapring layer}
As a coarse-to-fine manner to refine scene flow, the input coarse sparse flow in Fig.~\ref{figure:refinement} is first interpolated by Three-NN to obtain a coarse dense flow $\{sf_i^{dense} | ~i=1,\dots,n_1\}$. Then $PC_1=\{(x_i,pf_i) | ~i=1,\dots,n_1\}$ are warped by coarse dense flow to update the coordinates. The warped point clouds are denoted as $PC_1'=\{(x_i',pf_i) | ~i=1,\dots,n_1\}$ where $x_i' = x_i + sf_i^{dense}$.

\subsubsection{Attentive Cost Volume Layer}
\label{re-embedding}
\begin{figure}[t]
	\centering
	\includegraphics[width=1.00\linewidth]{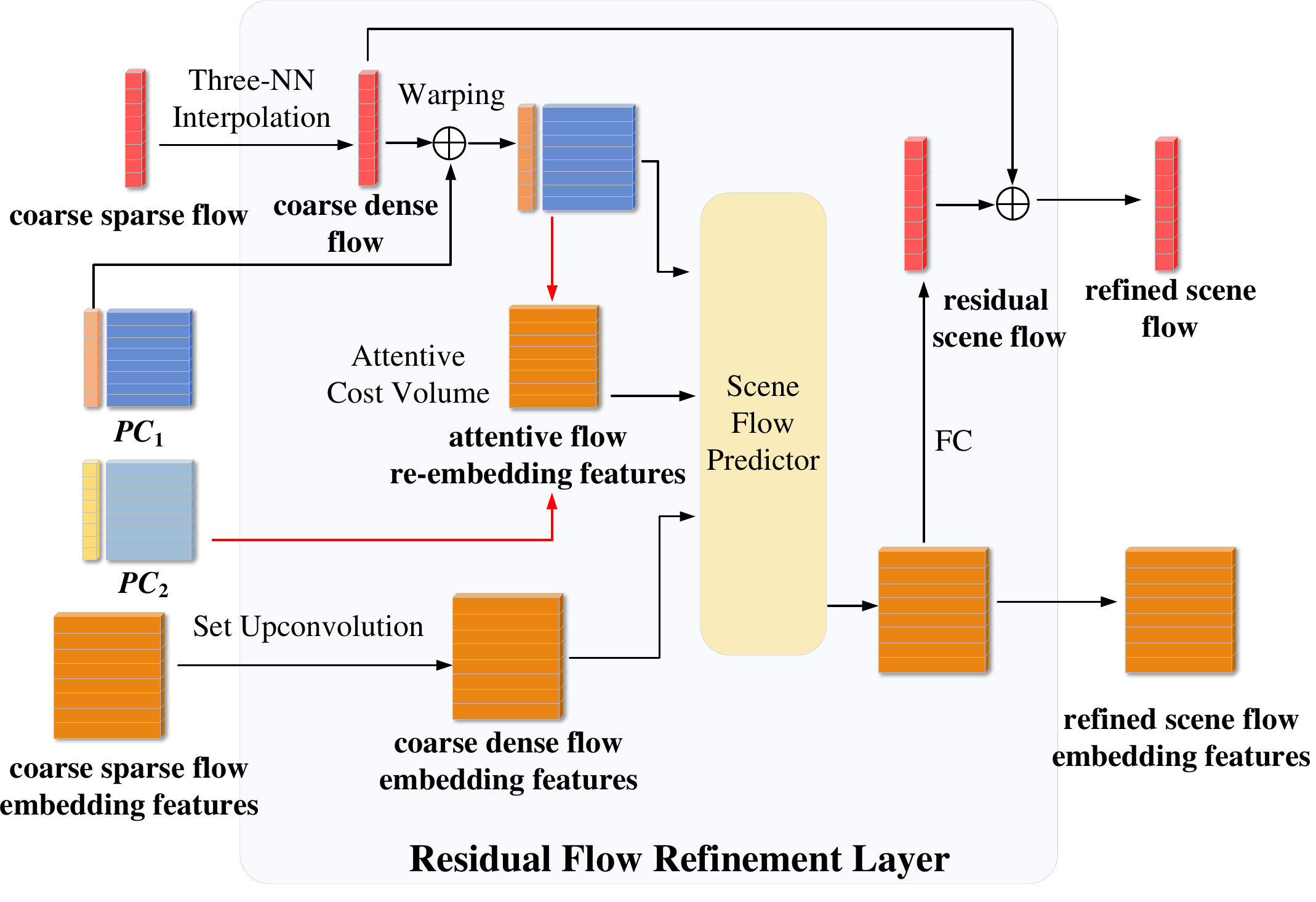}
	\vspace{-0.6cm}
	\caption{The details of residual flow refinement layer in Section \ref{section:refinement}.}   
	\label{figure:refinement}
\end{figure}

By taking in the warped point cloud $PC_1'$ in Section \ref{wapring layer} and the second frame of point cloud $PC_2$, the same layer of attentive cost volume described in Section \ref{section:context-aware Set Conv Layer} is adopted here to compute new flow embedding features $\{re_i |~ i=1,\dots,n_1 \}$. ``Attentive flow re-embedding features" is used in Fig.~\ref{figure:refinement} to emphasize its significance for scene flow predictor.

The skip connection in Fig.~\ref{figure:architecture} shows where $PC_1$ and $PC_2$ for each refinement layer come from. 
\subsubsection{Scene Flow Predictor}
The scene flow predictor is designed to refine the coarse dense flow embedding features. There are three inputs to this layer: point featurs of $PC_1$ $\{ pf_i \in \mathbb{R}^{c_1}|~ i=1,\dots,n_1 \}$, attentive flow re-embedding features $\{re_i \in \mathbb{R}^{c_2}|~ i=1,\dots,n_1 \}$ from Section \ref{re-embedding}, and the upsampled coarse dense flow embedding features $\{e_i \in \mathbb{R}^{c_3}|~ i=1,\dots,n_1 \}$ from Section \ref{set upconv}. The refined scene flow embedding features are calculated as:
\begin{equation}
	e_i' = MLP(pf_i \oplus re_i \oplus e_i).
\end{equation}

Instead of directly generating the refined scene flow by applying shared FC to the refined flow embedding features in \cite{wang2021hierarchical}, we exploit an explicit residual flow estimation structure. The residual scene flow $sf_i^{res}$ is first predicted by shared FC (implemented by $1\times1$ convolution) on the refined scene flow embedding features. Then, the coarse dense flow $sf_i^{dense}$ is added by the residual scene flow $sf_i^{res}$ to generate the refined scene flow $\{ sf_i |~ i=1,\dots,n_1 \}$:
\begin{equation}
	sf_i^{res} = FC(e_i'),
\end{equation}
\begin{equation}
	sf_i = sf_i^{res} + sf_i^{dense}.
\end{equation}

As shown in Fig.~\ref{figure:architecture}, the scene flow between two consecutive frames and scene flow embedding features are both refined through three residual flow refinement layers from coarse to fine. For the scene flow estimation of the finest level, we do not apply the residual flow refinement layer in order to save computational resources. Instead, we only use Three-NN interpolation and MLP to acquire the final residual scene flow and the overall scene flow.

Interpolation using three nearest neighbours from the preceding sparse layer is adopted to realize upsampling operation on flow embedding features while saving GPU memory. Then, MLP is used to predict final residual scene flow. Finally, the final residual scene flow is simply added to the interpolated coarse dense scene flow to obtain the overall 3D scene flow.

\section{Experiments}
\label{section:experiments}
\subsection{Datasets and Data Preprocessing}
We follow the supervised learning for scene flow estimation in our experiments. We first use the sythetic FlyingThings3D dataset \cite{mayer2016large} for both training and evaluation. Then, we also evaluate our trained model on real-world 3D scans from KITTI scene flow dataset \cite{menze2018object} without fine-tuning to demonstrate its generalization ability. 

Since the ground truth scene flow of large-scale real-world is difficult to acquire, we turn to the synthetic FlyingThings3D dataset \cite{mayer2016large} for our training. Following the preprocessing methods in \cite{gu2019hplflownet,wu2020pointpwc,wang2021hierarchical,li2021hcrf}, training and evaluation set are built by reconstructing ground truth scene flow and 3D point clouds from ground truth disparity map and optical flow. There are 19,640 pairs in training set and 3,824 pairs in the evaluation set. The points with depth larger than 35$m$ are excluded. 

KITTI Scene Flow 2015 dataset \cite{menze2018object} is another widely used dataset for scene flow estimation. It comprises 200 scenes for training and 200 scenes for testing. To evaluate our model and conduct a fair comparative experiment with other previous works, we recover the ground truth scene flow and point clouds following procedures in \cite{gu2019hplflownet,wu2020pointpwc,wang2021hierarchical,li2021hcrf}. Like FlyingThings3D \cite{mayer2016large}, points with depth larger than 35$m$ are also removed. The points from the ground with height less than 0.3$m$ are also removed since they are of no significance for flow estimation. 142 scenes from the training set with raw 3D point clouds are used for evaluation since ground truth scene flow is not available in the test set.

\setlength{\tabcolsep}{0.9mm}
\begin{table*}[!htbp]
	\begin{center}
		\caption{Comparison results between recent methods and ours on FlyingThings3D \cite{mayer2016large} and KITTI scene flow \cite{menze2018object} datasets. All listed approaches are only trained on FlyingThings3D dataset \cite{mayer2016large}. KITTI scene flow \cite{menze2018object} dataset is used to test the generalization ability of models. The best results are highlighted in bold. Weakly means weakly-supervised. Full means fully-supervised.}
		\label{table:flyingthing3d}
		\resizebox{1.00\textwidth}{!}
		{
			\begin{tabular}{clccccccccc}
				\toprule
				Evaluation Dataset       & Method      &  Training Data  &Input      & Sup. & EPE3D & Acc3D Strict         & Acc3D Relax & Outliers3D & EPE2D & Acc2D\\ \midrule
				& FlowNet3 \cite{ilg2018occlusions}   &  Quarter  & RGB stereo   & Full & 0.4570                        & 0.4179               & 0.6168                          & 0.6050   & 5.1348 &  0.8125                   \\ \cline{2-11}\noalign{\smallskip}
				& ICP \cite{besl1992method}   &  No    & Points & Full & 0.4062                        & 0.1614               & 0.3038                          & 0.8796   & 23.2280 & 0.2913                    \\
				& FlowNet3D \cite{liu2019flownet3d} &  Quarter  & Points & Full & 0.1136                        & 0.4125               & 0.7706                          & 0.6016      & 5.9740 & 0.5692                 \\
				& SPLATFlowNet \cite{su2018splatnet}&  Quarter & Points & Full  & 0.1205                        & 0.4197               & 0.7180                          & 0.6187     &  6.9759 & 0.5512                \\
				& HPLFlowNet \cite{gu2019hplflownet}   &   Quarter  & Points  & Full   & 0.0804                        & 0.6144               & 0.8555                          & 0.4287        & 4.6723 & 0.6764               \\
				& HPLFlowNet \cite{gu2019hplflownet}   &   Complete  & Points  &Full    &       0.0696                 &    ---          &      ---                    & ---     & --- &   ---          \\
				& PointPWC-Net \cite{wu2020pointpwc}   &   Complete  & Points  & Full   &       0.0588                &    0.7379          &             0.9276             & 0.3424     & 3.2390 &    0.7994         \\
				&  HALFlow \cite{wang2021hierarchical}      &  Quarter & Points & Full  & 0.0511                              &                  0.7808    &     0.9437                            &    0.3093   &    2.8739&         0.8056             \\ 
				&  HALFlow \cite{wang2021hierarchical}       &  Complete & Points  & Full &  0.0492                             &   0.7850                   &  0.9468                               &  0.3083      &2.7555 &            0.8111          \\
				&  FLOT \cite{puy2020flot}      &  Complete & Points & Full  & 0.0520                              &                  0.7320    &     0.9270                            &    0.3570   &    ---&         ---             \\
				\multirow{-9}{*}{\begin{tabular}[c]{@{}c@{}}FlyingThings 3D \\ dataset \cite{mayer2016large} \end{tabular}}
				
				&  HCRF-Flow \cite{li2021hcrf}      &  Quarter & Points & Full  & 0.0488                              &                  0.8337    &     0.9507                            &    0.2614   &    2.5652&       0.8704            \\
				& FlowStep3D  \cite{kittenplon2021flowstep3d}     &  Complete & Points & Full  & 0.0455                              &                 0.8162    &   0.9614                           &    0.2165  &    ---&     ---            \\
				&  Ours      &  Quarter & Points & Full & 0.0360                              &                  0.8894    &     0.9680                            &    0.1827   &    2.0418&         0.8888             \\ 
				&  Ours       &  Complete & Points  & Full &  \textbf{0.0310}                            &   \textbf{0.9139}                  & \textbf{0.9768}                               &  \textbf{0.1551}      &\textbf{1.7504} &          \textbf{0.9113}          \\
				\midrule
				& FlowNet3 \cite{ilg2018occlusions}    &   Quarter & RGB stereo  & Full  & 0.9111                        & 0.2039              & 0.3587                         & 0.7463     &  5.1023  & 0.7803                \\
				\cline{2-11}\noalign{\smallskip}
				& ICP \cite{besl1992method}   &  No    & Points &Full & 0.5181                        & 0.0669               & 0.1667                          & 0.8712     &  27.6752  & 0.1056                 \\
				& FlowNet3D \cite{liu2019flownet3d}  &   Quarter & Points & Full  & 0.1767                        & 0.3738               & 0.6677                          & 0.5271       &   7.2141  & 0.5093             \\
				& SPLATFlowNet \cite{su2018splatnet} &  Quarter & Points &Full & 0.1988                        & 0.2174               & 0.5391                          & 0.6575  &   8.2306  &  0.4189                  \\
				& HPLFlowNet \cite{gu2019hplflownet}  &   Quarter    & Points & Full  & 0.1169                        & 0.4783               & 0.7776                          & 0.4103      &   4.8055  &  0.5938                \\
				& HPLFlowNet \cite{gu2019hplflownet}  &  Complete   & Points  & Full  &        0.1113                &    ---             &       ---                    &  ---    &   ---   &     ---             \\
				& PointPWC-Net \cite{wu2020pointpwc}  &  Complete  & Points  &Full   &        0.0694               &  0.7281              &        0.8884                 & 0.2648   &    3.0062 &       0.7673        \\
				&  HALFlow \cite{wang2021hierarchical}     &  Quarter & Points & Full&  0.0692                             &   0.7532                  &   0.8943                              &   0.2529     &  2.8660 &    0.7811                  \\                                                                                                                                          &  HALFlow \cite{wang2021hierarchical}     &  Complete   & Points  & Full &  0.0622         &  0.7649 &         0.9026                        &      0.2492 &    2.5140 &  0.8128               \\ 
				&  FLOT \cite{puy2020flot}      &  Complete & Points & Full  & 0.0560                              &                  0.7550    &     0.9080                            &    0.2420   &    ---&         ---             \\
				\multirow{-9}{*}{\begin{tabular}[c]{@{}c@{}}KITTI \\ dataset \cite{menze2018object} \end{tabular}} 
				
				&  HCRF-Flow \cite{li2021hcrf}      &  Quarter & Points & Full  & 0.0531                              &                  0.8631    &     0.9444                           &    0.1797  &    2.0700&       0.8656            \\
				& FlowStep3D \cite{kittenplon2021flowstep3d}      &  Complete & Points & Full  & 0.0546                              &                  0.8051   &     0.9254                           &   \textbf{0.1492}  &    ---&       ---            \\
				&  Ours      &  Quarter & Points &Full & 0.0396                              &                  0.8679    &     0.9526                            &    0.1722   & 1.5517&         0.9125             \\ 
				&  Ours       &  Complete & Points & Full  &  \textbf{0.0351}                            &   \textbf{0.8932}                  & \textbf{0.9620}                               &  0.1654      & \textbf{1.2879} &          \textbf{0.9442}          \\
				\bottomrule
			\end{tabular}
		}
	\end{center}
\end{table*}

\setlength{\tabcolsep}{0.25mm}
\begin{table*}[!htbp]
	\begin{center}
		\caption{Ablation study on context-aware set convolution layer and residual flow learning.}
		\label{table:ablation study}
		\resizebox{1.00\textwidth}{!}
		{
			\begin{tabular}{c|ccccc|cc}
				\toprule
				Dataset   & Method               & EPE3D & Acc3D Strict         & Acc3D Relax & Outliers & EPE2D & Acc2D \\ \midrule
				& Baseline   &   0.0607
				&0.6977                      &     0.9349                          &  0.3366    &  3.3218  &     0.7591                \\
				&Baseline + Context-aware set convolution layer       &   0.0559
				&   0.7481                   &   0.9468                              &  0.3024      &  3.0917  &       0.7964                \\
				&Baseline + Residual flow learning       &   0.0384
				&   0.8757                   &   0.9637                              &  0.1996      &  2.1914  &       0.8765                \\

				\multirow{-4}{*}{\begin{tabular}[c]{@{}c@{}}FlyingThings 3D \\ dataset \cite{mayer2016large} \end{tabular}} & Ours (Baseline + Context-aware set convolution layer + Residual flow learning)    & \textbf{0.0360}                             &                  \textbf{0.8894}   &     \textbf{0.9680}                          &    \textbf{0.1827}   &   \textbf{2.0418} &         \textbf{0.8888}             \\  \bottomrule
			\end{tabular}
		}
	\end{center}
\end{table*}

\subsection{Training Details}

\subsubsection{Training Loss}
We train our network with a multi-scale supervision style, like \cite{sun2018pwc,wu2020pointpwc}. Assume the ground truth scene flow at $l$ level is ${SF^l} = \{{sf_i^l} \in \mathbb{R}^3|~ i=1,\dots,N_l\}$. The estimated scene flow at $l$ level is $\overline{SF^l} = \{\overline{sf_i^l} \in \mathbb{R}^3 |~i=1,\dots,N_l \}$. $N_l$ denotes the number of points at $l$ level. Then multi-scale loss is formulated as:
\begin{equation}
	Loss = \sum_{l=1}^4 \phi_l \frac{1}{N_l} \sum_{i=1}^{N_l} \left\|\overline{sf_i^l}-sf_i^l\right\|_2,
\end{equation}
where $\|\cdot\|_2$ denotes $L_2$ norm and $\phi_l$ is the weight of loss at $l$ level. We consider the last and finest scene flow as the $l=1$ level flow. In our network, the input point clouds have $4N=8192$ points and $N_1=N=2048$, $N_2=N/2=1024$, $N_3=N/8=256$, and $N_4=N/32=64$. The weights are set as $\phi_1=0.2$, $\phi_2=0.4$, $\phi_3=0.8$, and $\phi_4=1.6$.
\subsubsection{Implementation Details}
During the training and evaluation of our network, two consecutive frames of point clouds are randomly sampled to generate 8192 points as inputs, respectively. Following \cite{liu2019flownet3d,gu2019hplflownet,wu2020pointpwc,wang2021hierarchical,li2021hcrf}, the input point clouds of our network only contains 3D XYZ coordinates. $\frac{1}{4}$ of the training set (4910 pairs) of FlyingThings3D \cite{mayer2016large} dataset is first used to train our model and then complete training set is used to fine-tune our model in order to facilitate the training process. We conduct all the experiments on a single Titan RTX GPU with PyTorch 1.5.0. Adam optimizer is adopted in the training process with $\beta_1 = 0.9$ and $\beta_2=0.99$. The learning rate is initialized with 0.001 and exponentially decays with decay rate $\gamma = 0.5$. The step size for decaying is 80. The batchsize is set as 20.

\begin{figure*}[t]
	\centering
	\includegraphics[width=1.00\linewidth]{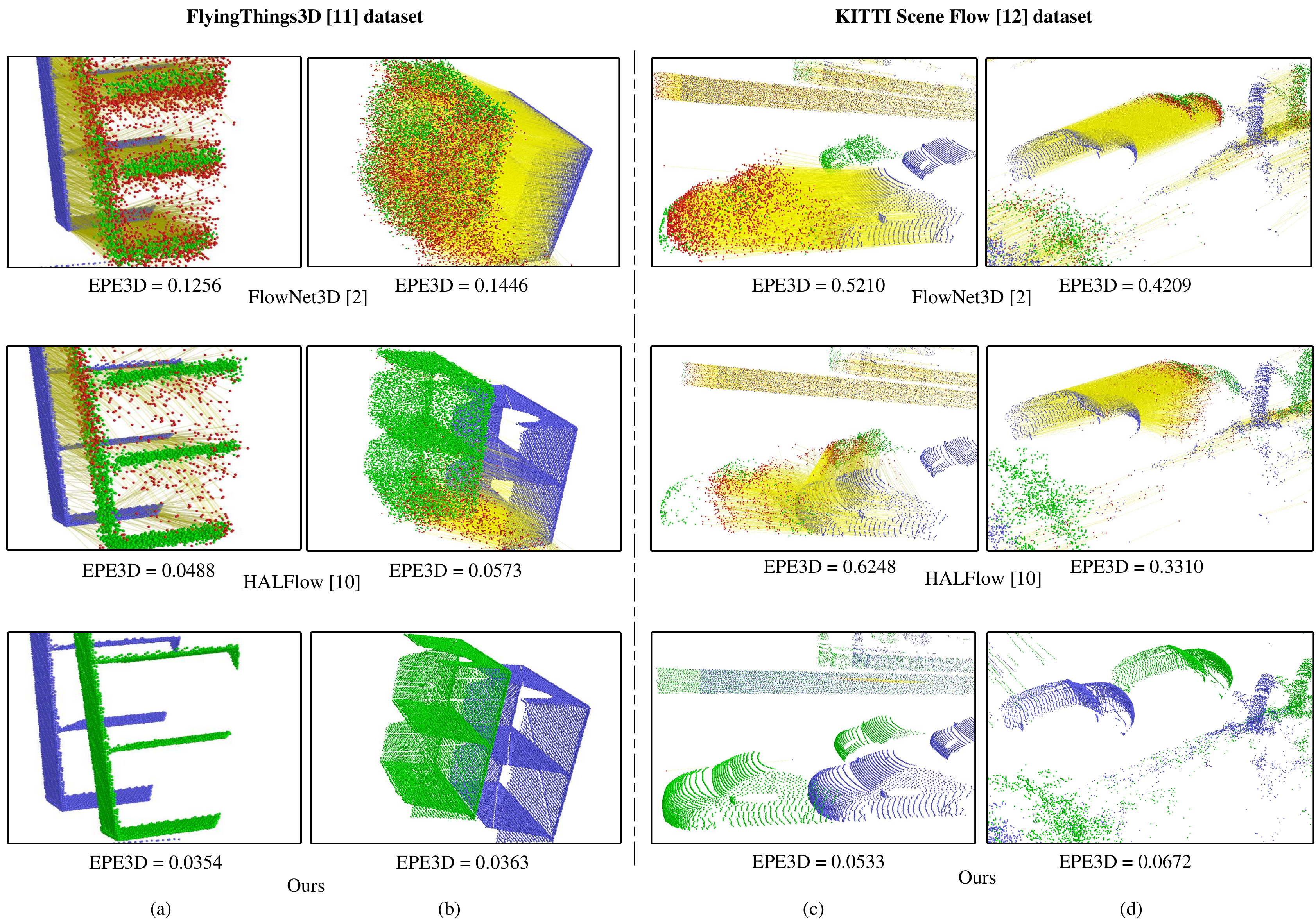}
	\vspace{-0.6cm}
	\caption{Detailed visualization comparison among FlowNet3D \cite{liu2019flownet3d}, HALFlow \cite{wang2021hierarchical}, and our method for scene flow estimation. Blue points are $PC_1$. Green points represent accurate predictions $\overline{PC_2}=PC_1+\overline{SF}$ and red points represent inaccurate predictions (Accuracy is measured by Acc3D Relax). Yellow lines connect inaccurately predicted red points in $\overline{PC_2}$ and their corresponding blue points in $PC_1$ to visualize the wrong predicted direction.}
	\label{figure:visualization}
\end{figure*}

\vspace{-0.06cm}
\subsection{Evaluation Metrics}
\vspace{-0.06cm}
The same evaluation metrics as \cite{liu2019flownet3d,gu2019hplflownet,wu2020pointpwc,wang2021hierarchical,li2021hcrf} are adopted to evaluate our model for a fair comparison with other methods. Let $sf_i$ be the ground truth scene flow and $\overline{sf_i}$ be the overall predicted scene flow. 

$\textbf{EPE3D}(m)$: $\frac{1}{N}\sum\limits_{i=1}^N \|\overline{sf_i}-sf_i\|_2$. 

$\textbf{Acc3D Strict}$: Percentage of $\overline{sf_i}$ such that $\|\overline{sf_i}-sf_i\|_2 < 0.05m$ or $\frac{\|\overline{sf_i}-sf_i\|_2}{\|sf_i\|_2} < 5\%$. 

$\textbf{Acc3D Relax}$: Percentage of $\overline{sf_i}$ such that $\|\overline{sf_i}-sf_i\|_2 < 0.1m$ or $\frac{\|\overline{sf_i}-sf_i\|_2}{\|sf_i\|_2} < 10\%$. 

$\textbf{Outliers3D}$: Percentage of $\overline{sf_i}$ such that $\|\overline{sf_i}-sf_i\|_2 > 0.3m$ or $\frac{\|\overline{sf_i}-sf_i\|_2}{\|sf_i\|_2} > 10\%$. 

$\textbf{EPE2D}(px)$:  $\frac{1}{N}\sum\limits_{i=1}^N \|\overline{of_i}-of_i\|_2$, where $of_i$ stands for the ground truth optical flow and $\overline{of_i}$ stands for the predicted optical flow from the projections of input point clouds and the point clouds synthesized by predicted 3D scene flow.

$\textbf{Acc2D}$: Percentage of $\overline{of_i}$ such that $\|\overline{of_i}-of_i\|_2 < 3px$ or $\frac{\|\overline{of_i}-of_i\|_2}{\|of_i\|_2} < 5\%$.

\subsection{Results}
\subsubsection{Comparison With State-of-the-Art (SOTA)}
\label{section:comparison}
Our method outperforms all other methods as quantitative evaluation results shown in Table \ref{table:flyingthing3d}. Our method surpasses SOTA method HCRF-Flow \cite{li2021hcrf} by 25\% on EPE3D on FlyingThings3D dataset, and achieves the best results on all 3D and 2D metrics. Our method also outperforms all other previous works remarkably on KITTI scene flow dataset \cite{menze2018object}, which strongly demonstrated the generalization ability of our method on real-world 3D data. Noticeably, recent SOTA methods \cite{li2021hcrf} impose rigid-body motion constraints while ours does not and outperforms them by a large margin.

Qualitative results in Fig.~\ref{figure:visualization} shows detailed visualization of the accuracy of the predicted scene flow by our method and other methods \cite{liu2019flownet3d,wang2021hierarchical}. For a structure with repetitive patterns like the multi-layer bookshelf shown in Fig.~\ref{figure:visualization}(a) and multiple cars shown in Fig.~\ref{figure:visualization}(c), the scene flow estimations are interfered by repetitive patterns for FlowNet3D \cite{liu2019flownet3d} and HALFlow \cite{wang2021hierarchical}. Points of $PC_1$ on a certain layer of the shelf learn the wrong correspondence with $PC_2$ due to similar and repetitive 3D structure. By comparison, our method introduces context-aware set convolution layer during feature extraction and performs better on repetitive patterns. The context-aware set convolution layer utilizes 3D coordinate information and relative position information from neighbourhood points together with feature information to extract context-aware point features. It allows the extracted local features to encode the correct global position within repetitive patterns in Euclidean space, providing more distinguishable positional cue for learning the right correspondence in the subsequent operation. For the scenes with long-distance motion on the right in Fig.~\ref{figure:visualization}(b)(d), our method performs the best among the three methods. We believe that it is the residual flow learning structure that contributes to the precise flow estimation. Explicit estimation of residual flow allows the coarse scene flow to be corrected with clear direction and compensates the errors successively by residual flow refinement layers.

\subsubsection{Ablation Study}
This paper proposes a novel context-aware set convolution layer and residual flow learning structure. In order to validate the effectiveness of each component, our network is trained and evaluated without either or both of the components using one quarter of the training set (4910 pairs). Other experiment settings are the same as Section \ref{section:comparison}. In Table \ref{table:ablation study}, the method without context-aware set convolution layer means removing the contextual weighting structure. Experiment results in Table \ref{table:ablation study} show that both of our proposed components improve the performance.

\section{Conclusion}
\label{section:conclusion}
In this paper, a novel context-aware feature encoding layer and a residual flow learning structure are proposed. Inspired by human attention on contextual perception and relocation by recognizing repetitive structure in space, our proposed context-aware set convolution layer aggregates point features utilizing 3D spatial structure and feature information for soft aggregation. Explicit residual flow learning is proposed to contribute to precise long-distance motion learning.
Ablation study is conducted to demonstrate their effectiveness for scene flow estimation.

Experiments on FlyingThings3D \cite{mayer2016large} and KITTI scene flow datasets \cite{menze2018object} shows that our method achieves state-of-the-art performance. We believe this context-aware feature extraction method and residual flow structure can improve the perception of repetitive patterns and long-distance motion estimation between two point clouds, which will provide some enlightenment for tasks with inter-frame association like registration, motion estimation, etc.

\ifCLASSOPTIONcaptionsoff
\newpage
\fi



%
\bibliographystyle{IEEEtran}  
\bibliography{IEEEabrv, root}


%
\vspace{-25pt}
\begin{IEEEbiography}[{\includegraphics[width=1in,height=1.25in,clip,keepaspectratio]{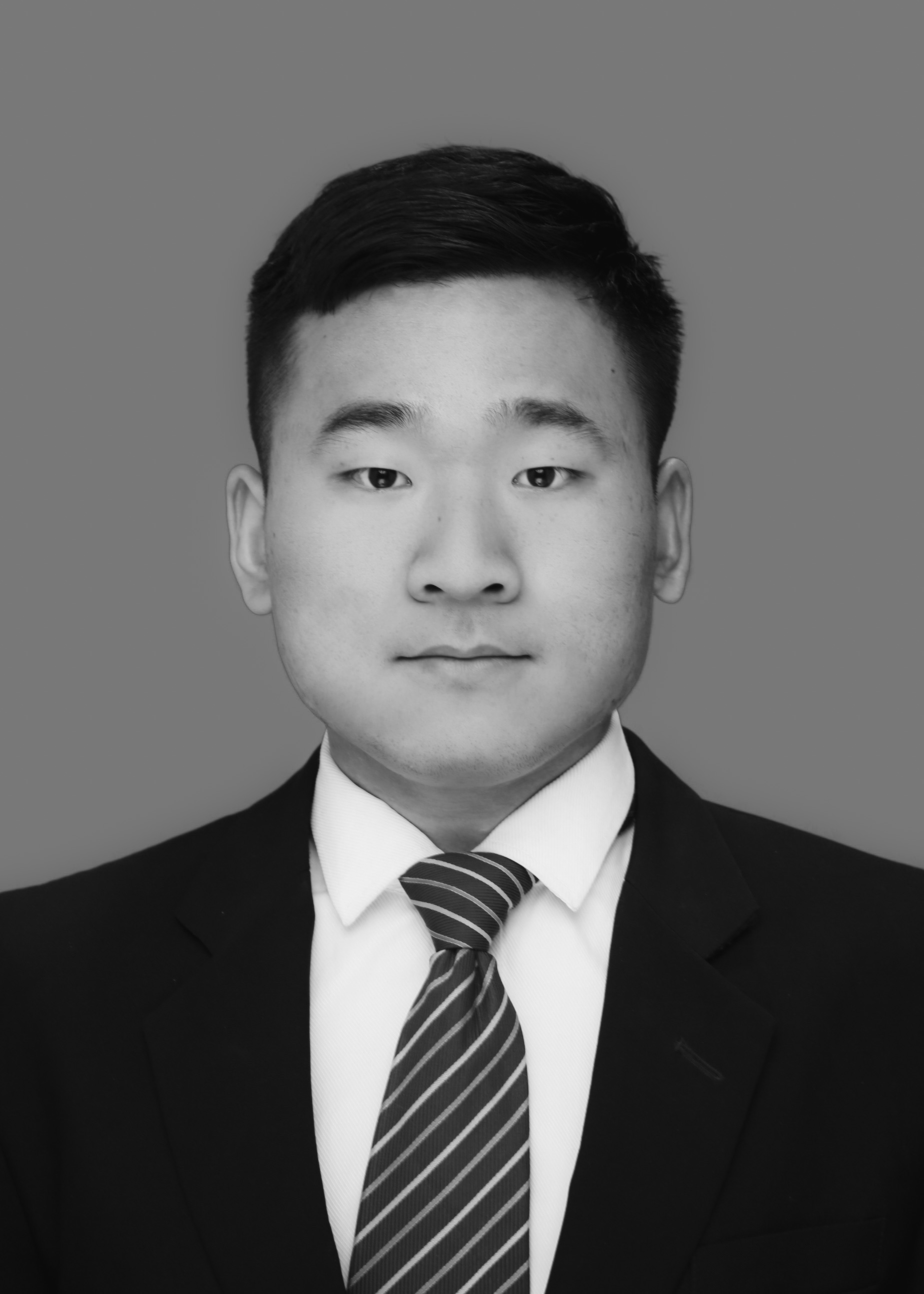}}]{Guangming Wang} received the B.S. degree from Department of Automation from Central South University, Changsha, China, in 2018. He is currently pursuing the Ph.D. degree in Control Science and Engineering with Shanghai Jiao Tong University. His current research interests include SLAM and computer vision, in particular, deep learning on point clouds.
\end{IEEEbiography}
\vspace{-25pt}
\begin{IEEEbiography}[{\includegraphics[width=1in,height=1.25in,clip,keepaspectratio]{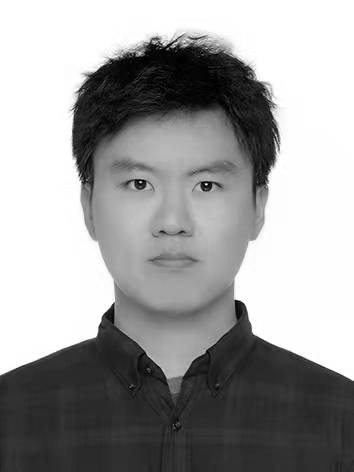}}]{YunZhe Hu} is currently pursuing the B.S. degree in Department of Automation, Shanghai Jiao Tong University. His latest research interests include 3D point clouds and computer vision.
\end{IEEEbiography}
\vspace{-25pt}
\begin{IEEEbiography}[{\includegraphics[width=0.9in,height=1.3in,clip]{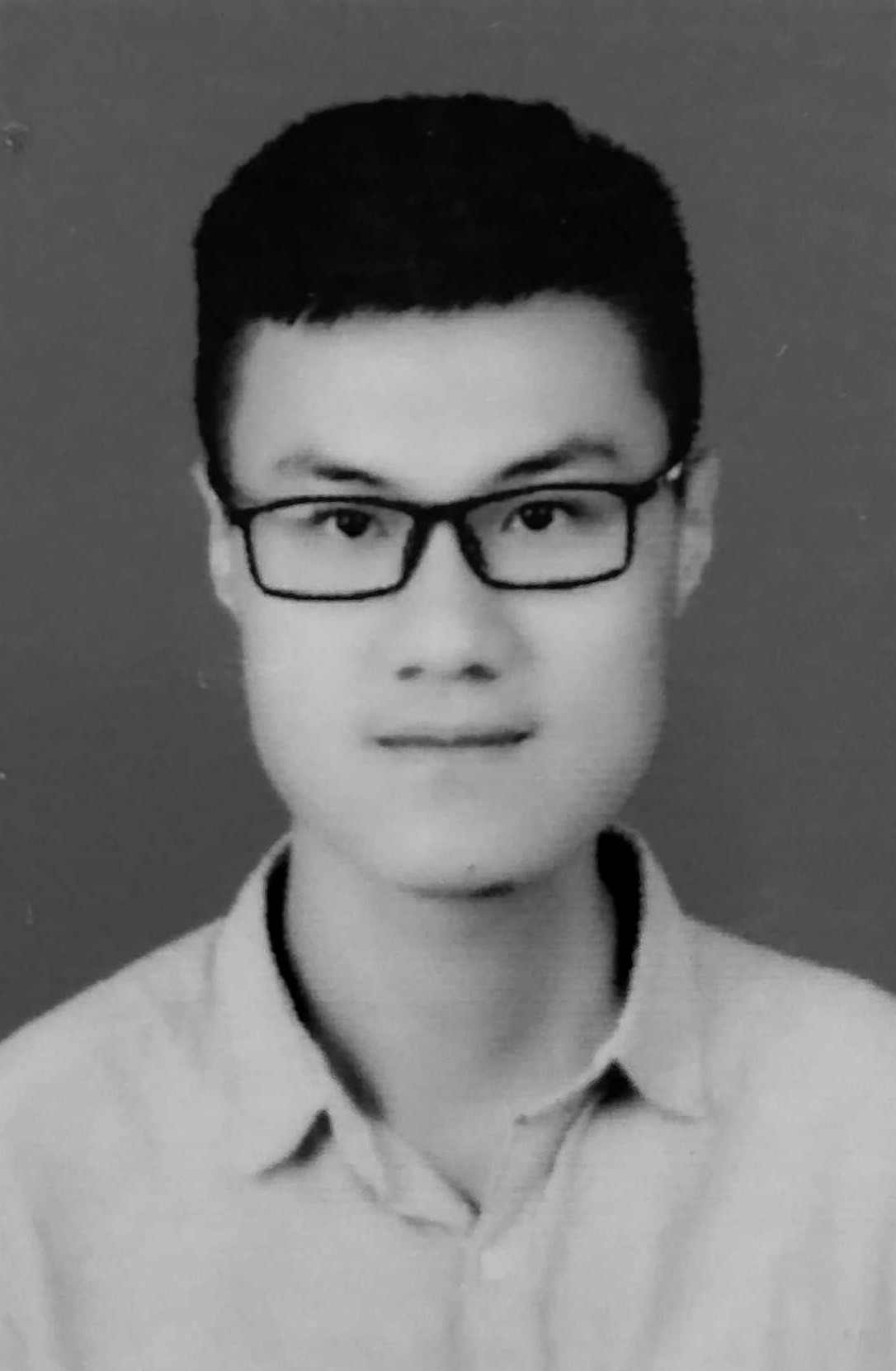}}]{Xinrui Wu} received the B.S. degree from the Department of Automation, Shanghai Jiao Tong University, Shanghai, China, in 2021, where he is currently pursuing the M.S. degree in Control Science and Engineering. His latest research interests include 3D point clouds and computer vision.
\end{IEEEbiography}
\vspace{-25pt}
\begin{IEEEbiography}[{\includegraphics[width=1in,height=1.25in,clip,keepaspectratio]{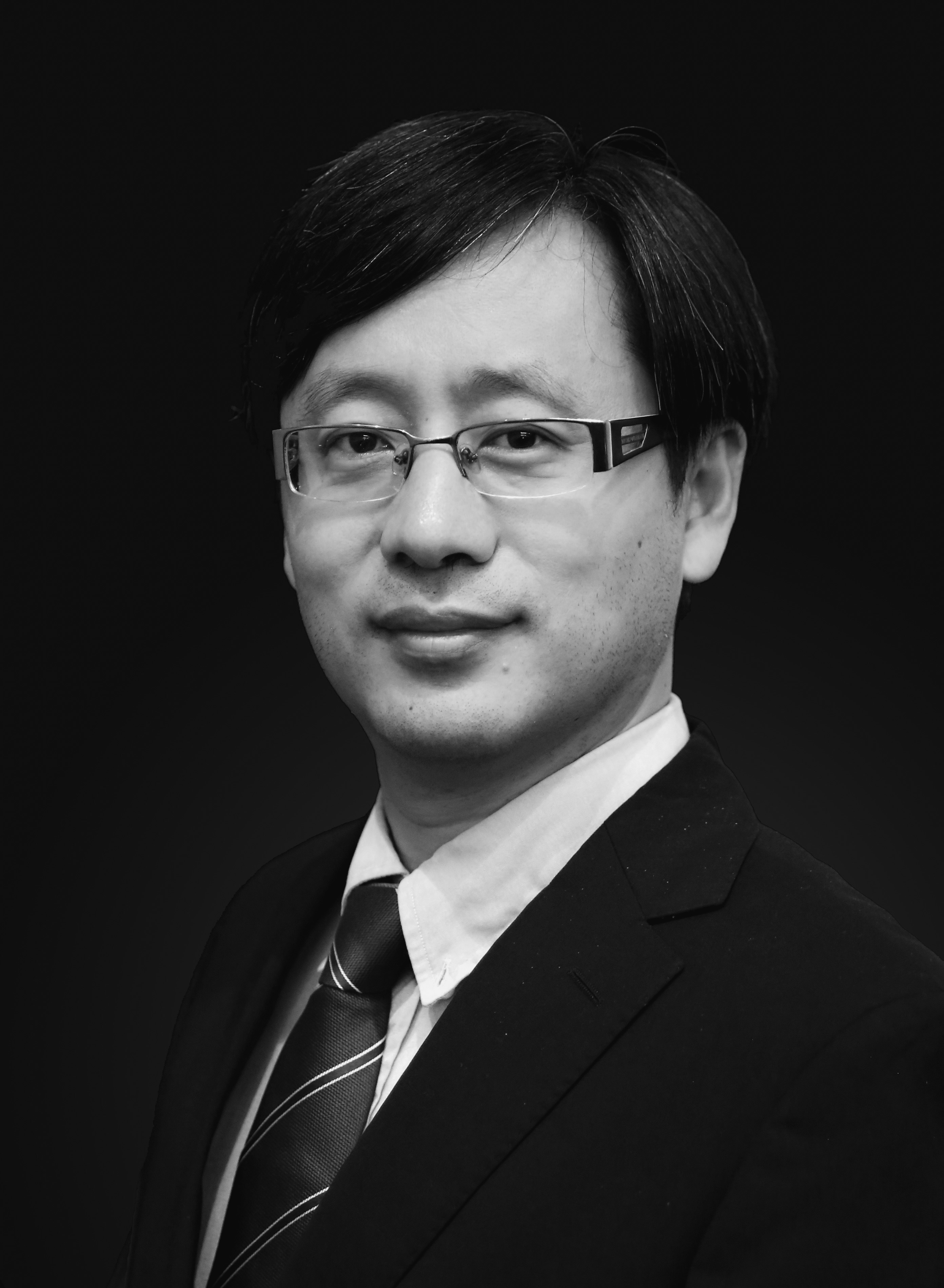}}]{Hesheng Wang} (SM'15) received the B.Eng. degree in electrical engineering from the Harbin Institute of Technology, Harbin, China, in 2002, and the M.Phil. and Ph.D. degrees in automation and computer-aided engineering from The Chinese University of Hong Kong, Hong Kong, in 2004 and 2007, respectively. He is currently a Professor with the Department of Automation, Shanghai Jiao Tong University, Shanghai, China. His current research interests include visual servoing, service robot, computer vision, and autonomous driving. Dr. Wang is an Associate Editor of IEEE Transactions on Automation Science and Engineering, IEEE Robotics and Automation Letters, Assembly Automation and the International Journal of Humanoid Robotics, a Technical Editor of the IEEE/ASME Transactions on Mechatronics, an Editor of Conference Editorial Board (CEB) of IEEE Robotics and Automation Society. He served as an Associate Editor of the IEEE Transactions on Robotics from 2015 to 2019. He was the General Chair of the IEEE RCAR 2016, and the Program Chair of the IEEE ROBIO 2014 and IEEE/ASME AIM 2019. He is the General Chair of IEEE ROBIO 2022.
\end{IEEEbiography}




\end{document}